\documentclass{article}
\write18{}
\usepackage{amssymb}
\usepackage{amsmath}
\usepackage{algorithm}
\usepackage{booktabs}
\usepackage{multirow}
\usepackage{algpseudocode}
\usepackage{graphicx}
\usepackage{bbm}
\usepackage{caption}
\usepackage{subcaption}
\usepackage{xcolor}
\usepackage{lipsum}
\usepackage{svg}
\usepackage{siunitx}
\usepackage{spconf,amsmath,graphicx}
\usepackage{enumitem}
\usepackage{url}

\title{SuperCM: Revisiting Clustering for Semi-Supervised Learning}
\name{Durgesh Singh,  Ahc\`ene Boubekki,  Robert Jenssen,  Michael C. Kampffmeyer\thanks{Visual Intelligence publications are financially supported by the Research Council of Norway, through its Centre for Research-based Innovation funding scheme (grant no. 309439), and Consortium Partners. The authors are part of the UiT Machine Learning Group: https://machine-learning.uit.no.}}
\address{Department of Physics and Technology, UiT The Arctic University of Norway, Tromsø }

\begin{document}
%
\maketitle
%

\begin{abstract}

The development of semi-supervised learning (SSL) has in recent years largely focused on the development of new consistency regularization or entropy minimization approaches, often resulting in models with complex training strategies to obtain the desired results. In this work, we instead propose a novel approach that explicitly incorporates the underlying clustering assumption in SSL through extending a recently proposed differentiable clustering module. 
Leveraging annotated data to guide the cluster centroids results in a simple end-to-end trainable deep SSL approach. We demonstrate that the proposed model improves the performance over the supervised-only baseline and show that our framework can be used in conjunction with other SSL methods to further boost their performance.


\end{abstract}
\section{Introduction}
\label{sec:intro}
Traditional deep learning has achieved state-of-the-art performance on various tasks at the cost of large-scale supervised training data.  However, it is difficult to obtain such a dataset in many applications, due to an expensive and time-consuming annotation process~\cite{Willemink2020}. Several approaches have tackled this dependency by exploiting information from the otherwise abundant unlabeled data, and thereby improving the existing model performance. Semi-supervised Learning (SSL)~\cite{Chapelle2006} is one such paradigm that addresses the problem of label scarcity. SSL methods depend on the clustering assumption~\cite{Yang2021ASO} and mostly leverage consistency regularization ~\cite{Tarvainen2017, Laine2017TemporalEF, Miyato2019VirtualAT,Athiwaratkun2019ThereAM} or entropy minimization~\cite{Grandvalet2005,Dong-HyunLee2013,Pham2021MetaPL,Zhai2019S4LSS} to enforce the same. However, the success of  consistency regularization and entropy minimization methods depends on the choice of an appropriate perturbation strategy or the quality of the estimated pseudo-labels, respectively. 
This has led to complex training mechanisms that incorporate different perturbation strategies and rely on various  pseudo-label heuristics to achieve greater performance~\cite{Miyato2019VirtualAT,Zhai2019S4LSS}. 
 
\begin{figure}[htb]
  \centering
  \begin{minipage}[b]{\linewidth}
  \centering
  \centerline{(a) Cross-Entropy \hspace{6em} (b) SuperCM}   
  \includegraphics[width=\linewidth]{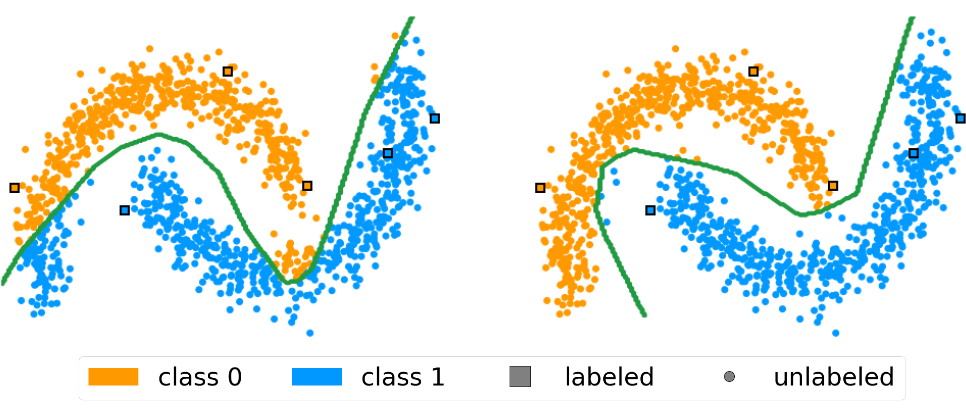}
\end{minipage}
\vspace{-0.8cm}
\caption{Comparison of Cross-Entropy and SuperCM on the two moons dataset for 3 labeled examples per class out of a total of 1600 datapoints. Each approach leverages a feature extractor consisting of a fully connected neural network with three hidden layers, each with 10 neurons.}
\vspace{-0.35cm}
\label{fig:sslmoons}
\end{figure}

\begin{figure*}[!t]
 \centering
 \includegraphics[width=0.95\textwidth]{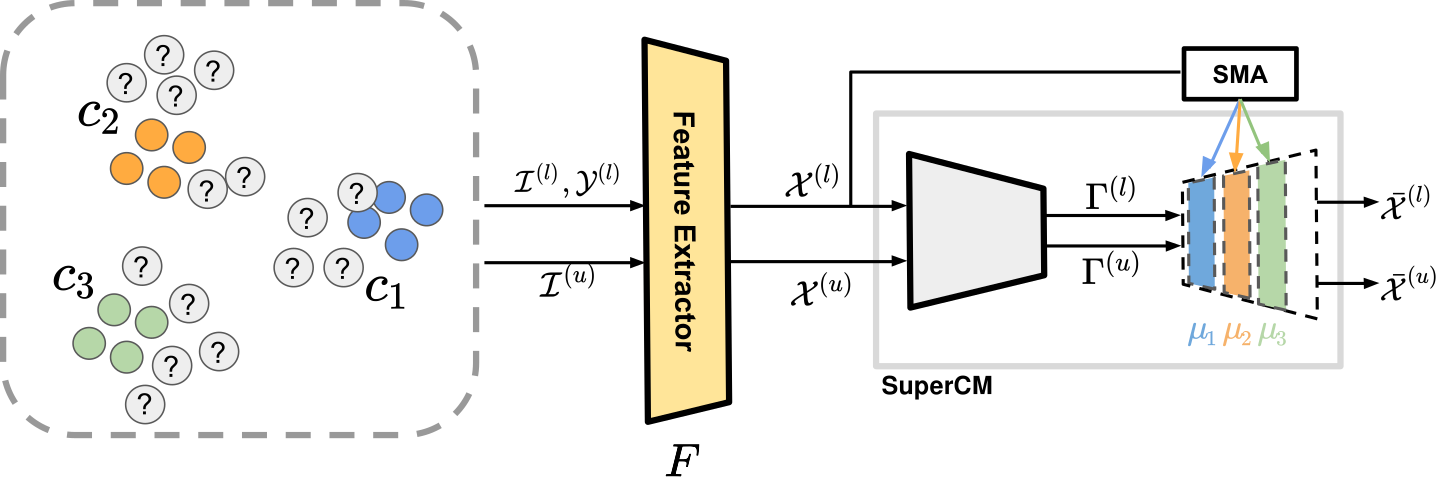}
 \vspace{-0.2cm}
 \caption{Architecture of the SuperCM. 
 The feature extractor and the CM's encoder are trained with gradient descent using both labeled (colored) and unlabeled (gray) data. The centroids of the CM are updated as the class-wise moving average (MA) of the labeled data. }
  \vspace{-0.4cm}
 \label{fig:supercm}
\end{figure*}

In this work, we take a different approach and study the problem of SSL from the clustering perspective. Recent deep clustering approaches partition high dimension input features by performing clustering  and representation learning simultaneously~\cite{Zhou2022}. Owing to the recent success of these methods, we develop a novel training approach that takes inspiration from the Gaussian mixture model (GMM), and utilizes a one layer auto-encoder called the clustering module (CM)~\cite{Boubekki2021JointOO} for the SSL task. Our method is called Semi-\textbf{Super}vised \textbf{C}lustering \textbf{M}odule (SuperCM), which does not rely on a complex training scheme and achieves performance improvement with respect to its supervised-only baseline. 
As an illustration, we show in Figure~\ref{fig:sslmoons}, how SuperCM compares to a model trained with a vanilla Cross-Entropy (CE), and how it results in better separation of the two moons dataset.
The SuperCM, due to the differentiable nature of the CM, can be further incorporated as a regularizer in other gradient-based SSL methods to boost their performance, thus opening new avenues in clustering-based SSL research.  
We summarize the contribution of our work as follows:
\begin{itemize}[leftmargin=*]
    \item[-] We develop a simple approach that builds on a differentiable clustering module and explicitly enforces the clustering assumption in the SSL task.
   \item[-] Our SuperCM is complimentary to other SSL methods and improves their performance significantly when the number of annotated samples is low.
\end{itemize}

\section{Related Works}
\label{sec:relatedworks}
To set the stage for SuperCM, we discuss relevant deep SSL and clustering approaches. For a more extensive survey the interested reader is referred to~\cite{Yang2021ASO,Zhou2022}.

\subsection{Semi-Supervised Learning}
\label{sec:ssl}
We first discuss representative methods based on consistency regularization and entropy minimization for the SSL task. Consistency regularization methods encourage invariant predictions for different input perturbations of the same unlabeled data point. 
One prominent example of consistency-based methods is Virtual Adversarial Training (VAT)~\cite{Miyato2019VirtualAT}, which uses the concept of adversarial attacks for consistency regularization. It aims to find a perturbation to the input data in an adversarial direction, and enforces the consistency between the model predictions for the original input and its perturbation. Moreover,
entropy minimization~\cite{Grandvalet2005} encourages low entropy on the model predictions for the unlabeled data. One representative method of entropy minimization is Pseudo-label~\cite{Dong-HyunLee2013} that generates high-confidence proxy labels for the unlabeled data to guide the training. While these approaches have shown performance improvement in the supervised-only baseline, they rely on complex training strategies to achieve the same.

\subsection{Deep Clustering}
In this section, we discuss deep clustering approaches~\cite{Zhou2022}, which aim to cluster high dimensional data with the help of neural networks. Early methods~\cite{Xie2016UnsupervisedDE, Caron2018DeepCF, Li2021PrototypicalCL} perform iterative training by minimizing the clustering loss for learning deep features and using the updated features to estimate new cluster labels. However, this alternating training strategy can be sub-optimal. Recent approaches perform joint representation learning and clustering simultaneously to obtain more cluster-friendly embedding. One prominent approach of simultaneous clustering is the CM~\cite{Boubekki2021JointOO}, which uses a one-layer auto-encoder with a deep feature extractor and performs clustering by minimizing a GMM-based clustering loss along with a suitable representation learning loss.
We briefly describe the CM in the next section and discuss how it can be extended to the learning in the SSL scenario.

\begin{table*}
\centering
\caption{Results for Top-1 accuracy with CE and  SuperCM for different SSL base models on the CIFAR-10 dataset. None denotes the SSL setting without a base model where performance of CE and SuperCM is compared. Bold numbers indicate statistically significant improvements (t-test, $p < 0.05$).
}
\begin{tabular*}{.99\linewidth}{@{\extracolsep{\fill}}lcccc}
\toprule
                                               & \multicolumn{2}{c}{600 labels}              & \multicolumn{2}{c}{4000 labels}    \\ 
                                SSL base model               & CE                         & SuperCM                    & CE              & SuperCM                             \\ \midrule
None                                  & 56.94$\pm$0.46  & \textbf{62.14$\pm$1.40}               & 78.65$\pm$0.45            & \textbf{82.26$\pm$0.26}    \\ 
Pseudo-Label~\cite{Dong-HyunLee2013}  & 61.05 $\pm$1.25            & \textbf{65.19 $\pm$2.52}   & 84.97$\pm$0.22             & 85.19 $\pm$0.47  \\
VAT~\cite{Miyato2019VirtualAT}         & 68.43 $\pm$0.89            & \textbf{75.23 $\pm$3.92}   & 86.82 $\pm$0.19   & 86.69 $\pm$0.11            \\
\bottomrule
\end{tabular*}
\label{tab:res-reg}
\vspace{-0.4cm}
\end{table*}

\section{Method}
\label{sec:method}

\subsection{Clustering Module}

As the key building block of our SSL approach, we first describe the CM introduced in~\cite{Boubekki2021JointOO}. The model aims to maximize a differentiable, rephrased version of the $\mathcal{Q}$-function of a Gaussian mixture model. The loss function of the CM can be stated as follows: 
\begin{multline}
\mathcal{L}_{\rm CM} = \! \!\frac{1}{N}  
\Bigg( \! \sum_{i=1}^N ||{\pmb x}_{i} - \bar{{\pmb x}}_{i} ||^2 \!  
+ \sum_{i=1}^N \sum_{k=1}^K  \gamma_{ik} ( 1 -\gamma_{ik}) || {\pmb \mu}_k||^2 \\ 
- \sum_{i=1}^N\sum_{ \substack{k,l=1 \\ k \neq l}  }^K \gamma_{ik} \gamma_{il } {{\pmb \mu}_k^T} {\pmb \mu}_l
  +  \sum_{k=1}^K (1- \alpha_k) \log \tilde{\gamma}_k \Bigg)
 \label{eq:qfunc}
\end{multline}
where  $N>0$ is the number of data points and $K>0$ is the number of clusters. An input data ${\pmb x}_{i} \in \mathcal{X} \subset \mathbbm{R}^d $ has a label $y_{i} \in \mathcal{Y}$ and its reconstruction by the CM is denoted $\bar{{\pmb x}}_i \in \mathbbm{R}^d$. 
The CM's encoder, with softmax activation, outputs the set $\Gamma = \{ \gamma_{ik}=P(y_{i} = k|{\pmb x}_{i})\}$ which corresponds to the posterior probabilities of the cluster assignment, also known as cluster responsibilities. 
The centroids of the model, denoted ${\pmb \mu}_{k} \in R^d$, are the weights of the CM's decoder.
Finally, the cluster probabilities are controlled by a Dirichlet prior over the $\tilde\gamma_{k} = \mathbbm{E}_{\mathcal{X}}(\gamma_{ik})$ with concentration $\alpha>1$.



\subsection{SuperCM} 
\label{sec:supercm}
An intuitive SSL extension of the CM would be to prepend a feed-forward neural network to the CM module and leverage the supervised data through the CE loss over the posterior probabilities. However, we observe that this could lead to trivial solutions as the training of the backbone is too fast.
We overcome this obstacle by learning the centroids as a class-wise average of the labeled data instead of gradient descent. Specifically, we rely on a moving average to prevent frequent noisy updates. 
Computing the centroids this way also prevents them from collapsing and makes the Dirichlet prior along with its hyper-parameter $\alpha$ unnecessary.
We call this approach SuperCM and show a schematic representation of the model in Figure~\ref{fig:supercm}.

At each iteration $t$, the input consists of $n^{(l)}$ labeled data pairs $\{\mathcal{I}^{(l)},\mathcal{Y}^{(l)}\}$ and $n^{(u)}$ unlabeled data $\mathcal{I}^{(u)}$.
Both inputs are first transformed by a feature extractor $F$ into
  $\mathcal{X}^{(l)}=F\big(\mathcal{I}^{(l)}\big)$ and $\mathcal{X}^{(u)}=F\big(\mathcal{I}^{(u)}\big)$.
The centroids are then updated using the labeled data as follows:
\begin{equation}
    {\pmb \mu}_{k} = \frac{t-1}{t} {\pmb \mu}_{k} + \frac{1}{t} \frac{1}{n^{(l)}} \sum_{i=1}^{n^{(l)}} \mathbbm{1}_{(y^{(l)}_{i}=k)} {\pmb x}^{(l)}_i
\end{equation}
Finally, the labeled and unlabeled data are concatenated and passed through the CM to obtain the cluster probabilities $\Gamma^{(l+u)}$ as well as the reconstructions $\bar{\mathcal{X}}_{l+u}$.

The loss function of SuperCM combines a standard CE loss applied on the cluster responsibilities of the labeled data with the CM loss applied on both types of data. 
The SuperCM can also be used as a regularizer for existing SSL models. The combined loss can be stated as follows:
\begin{align}
\begin{split}
    \mathcal{L}^{\rm{SSL}}_{\rm SuperCM} & = 
        \mathbf{CE}( \Gamma^{(l)}, \mathcal{Y}^{(l)} ) \\
        & + \beta \cdot \mathcal{L}_{\rm CM}( \mathcal{X}^{(l+u)}, \Gamma^{(l+u)}, \bar{\mathcal{X}}^{(l+u)}) \\
        & + \delta \cdot \mathcal{L}_{\rm SSL}
 \label{eq:supcm}
\end{split}
\end{align}
where  $\beta \geq 0$ and $\delta \geq 0$ are weights of the CM loss and of the loss of the SSL base model, respectively\footnote{ Code available at https://github.com/Durgesh93/SuperCM.git}.

\section{Experiments}
\label{sec:experiments}
In this section, we compare the performance of the SuperCM as an SSL model, and as a regularizer for other SSL baselines.
We follow the recommendations of~\cite{Oliver2018} for data pre-possessing, model architecture, and training protocol.   

\subsection{Experimental Setting}

\textbf{Data} We use CIFAR-10~\cite{cifar10} for all the experiments. It consists of 60000, $32 \times 32$ color images distributed into ten classes. 
We use data augmentation random-crop, flip and Gaussian noise.
The dataset is divided into training, validation, and test sets containing 50000, 5000, and 10000 images, respectively.  \\
\textbf{Architecture} We use the Wide-ResNet-28-2~\cite{Zagoruyko2016WideRN} architecture with 1.5M parameters as backbone for all the models.  The architecture returns feature vectors before the linear classifier of dimension $128$.
When SuperCM is involved, the classifier is replaced with the CM's one-layer autoencoder. \\
\textbf{Training} In our experiments, the model is trained using the Adam~\cite{Kingma2015AdamAM} optimizer for 500000 iterations with batch size 100. The learning rate is decayed once with a factor of 0.1 after 400000 iterations. 
The hyper-parameters $\beta$ and $\delta$ are tuned over the validation dataset. \\
\textbf{Baseline} There are two types of baselines for our experiments. For the SSL setting, the baseline is supervised-only training with the CE loss, i.e. $\delta=0$ and $\beta=0$ in Eq.~\eqref{eq:supcm}. For the SSL regularization setting, we use VAT and Pseudo-label as base models, i.e. $\delta>0$ and $\beta=0$ in Eq.~\eqref{eq:supcm}.\\
\textbf{Evaluation} The final model is computed based on stochastic weight averaging~\cite{Izmailov2018AveragingWL} for all the experiments.
Model selection is based on the best Top-1 accuracy on the validation set.
Overall model performance is measured using Top-1 accuracy on the test set. We report mean and standard deviation over five runs trained with different random seeds.

\subsection{Results}

Table~\ref{tab:res-reg}  summarizes the results for the 600 and 4000 label settings on the CIFAR-10 dataset. 
Without the SSL base model, i.e., $\delta=0$, SuperCM significantly improves the performance of the CE baseline by $5.2\%$ and $3.6\%$ for 600 and 4000 labels respectively in the SSL setting.
When SuperCM is used as a regularizer of a base model, the accuracy of the base model significantly increases for the 600 labels setting.
Specifically, SuperCM improves the accuracy of the Pseudo-Label baseline by $4.14\%$ and that of VAT by $6.8\%$.
However, we do not observe significant improvement for the 4000 labels setting. 
Our hypothesis for the results in the low supervision setting is that- the SSL base models benefit from the well-separated clustering obtained by SuperCM, when it is difficult to obtain a reliable supervisory signal with the CE loss in the training. A visualization of the well-separated features learned by SuperCM is provided in Sec.~\ref{sec:featureVis}.



\section{Analysis}
\label{sec:analysis}
In this section, we analyze different aspects of the SuperCM and discuss the results.

\subsection{Feature Visualization}
\label{sec:featureVis}
\begin{figure}[!t]
  \centering
\begin{minipage}[b]{.48\linewidth}
  \centering
  \centerline{(a) CE}\medskip
  \includegraphics[width=\linewidth]{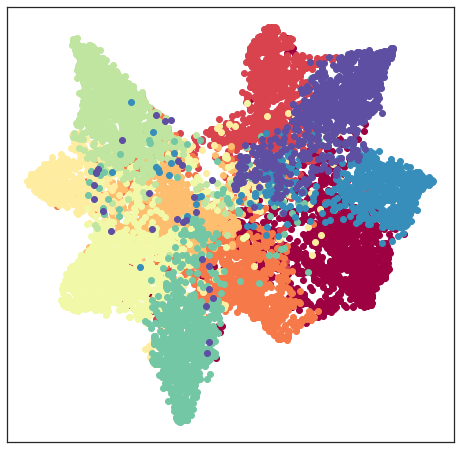}
  
\end{minipage}
\begin{minipage}[b]{0.48\linewidth}
  \centering
  \centerline{(b) SuperCM}\medskip
  \includegraphics[width=\linewidth]{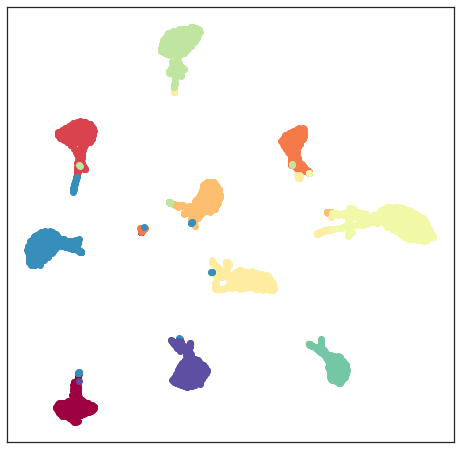}
\end{minipage}
\vspace{-0.5cm}
\caption{UMAP plots for backbone features of the models trained with CE and SuperCM.}
\vspace{-0.4cm}
\label{fig:umap}
\end{figure}

We show in Figure~\ref{fig:umap}, the UMAP\cite{McInnes2018UMAPUM} representations of the feature space learned by the models trained with CE and SuperCM for the CIFAR-10 600 labels configuration.
It is evident that SuperCM yields more separated and compact classes, which leads to better generalization and thereby better performance. 

\subsection{Varying Data Amounts}
\label{sec:vardata}
\begin{figure}[!t]
\begin{minipage}[b]{\linewidth}
 \centering
 \centerline{ \includegraphics[width=\textwidth]{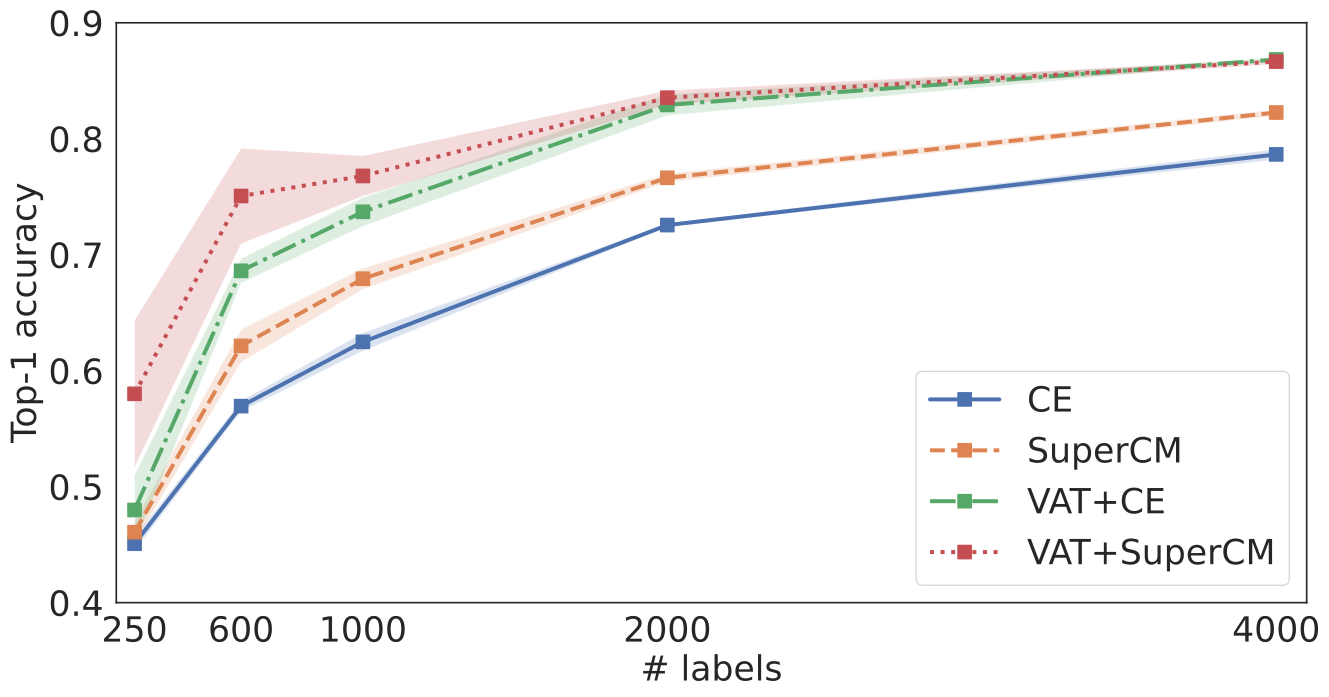}}
\end{minipage}
\vspace{-0.9cm}
\caption{Training with CE and SuperCM for different levels of supervision.}
\vspace{-0.4cm}
\label{fig:vary}
\end{figure}

We evaluate the performance of our method on CIFAR-10 for different training label sizes ranging from 250 to 4000 labels, with and without  VAT as the SSL base model. The results are summarized in Figure~\ref{fig:vary}.


Without the SSL base model, SuperCM improves the supervised-only baseline (CE) for all levels of supervision. However, we observe that improvements diminish as the number of labels drops significantly.
In this case, we do not expect a large performance improvement from SuperCM alone, as the supervision is extremely scarce and not sufficient to learn a cluster-friendly embedding for the overarching SSL task. However, as the number of labels increases, we see significant improvements compared to the supervised-only baseline ranging from $3\%$ to $6\%$. 

Similarly, incorporating SuperCM as a regularizer improves the performance of VAT significantly for all except the 2000 and 4000 label settings. It is interesting to observe that SuperCM regularization improves the VAT baseline by around $10\%$ in the 250 label setting, where VAT benefits from the online clustering regularization of the SuperCM. We also see performance improvements of around $6\%$  and $3\%$ in the case of the 600 and 1000 label settings.

These results highlight the benefit of our method without relying on a complex training strategy, unlike other prominent approaches in the SSL research.  

\subsection{Hyper-parameter Sensitivity}
\begin{figure}[!t]
\begin{minipage}[b]{\linewidth}
\centering
\centerline{\includegraphics[width=\textwidth]{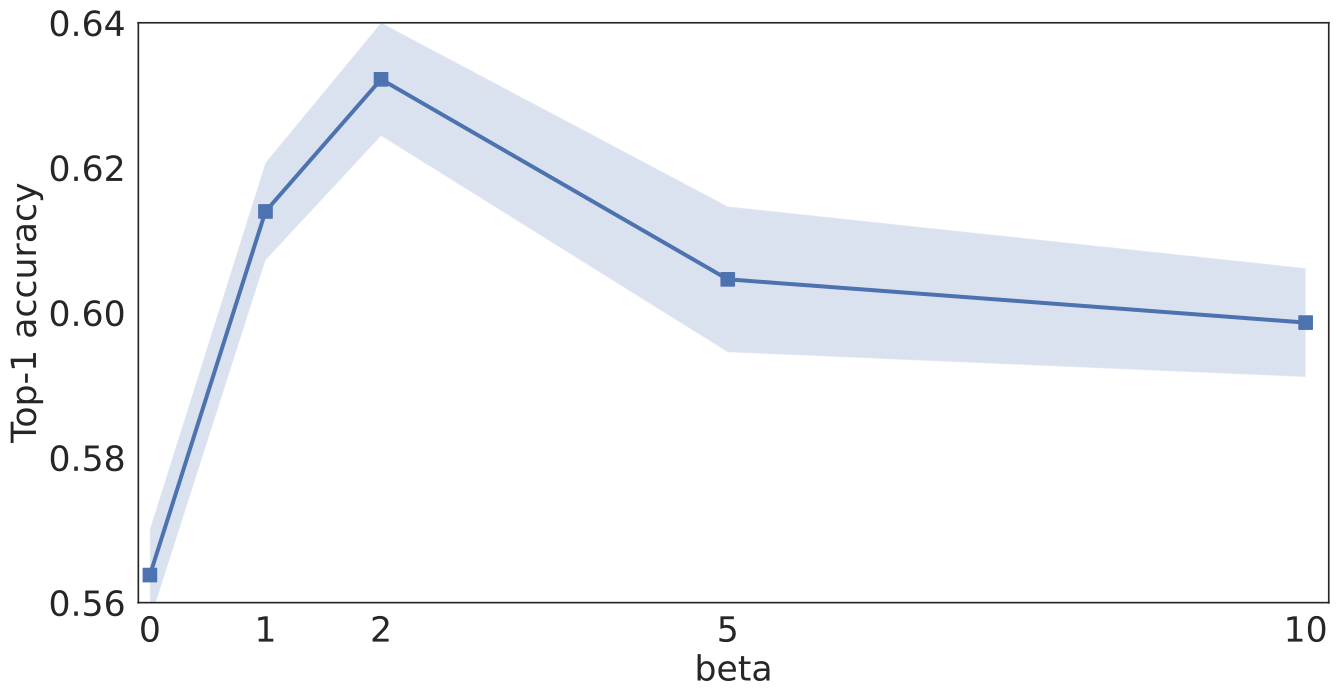}}
\vspace{-0.5cm}
\caption{Top-1 accuracy of SuperCM trained with 600 labels and different values of the hyper-parameter $\beta$.}
\label{fig:paramsen}
\vspace{-0.4cm}
\end{minipage}
\end{figure}

Without a base model, the SuperCM uses a single hyper-parameter $\beta$. 
To study the sensitivity of the model toward $\beta$, we trained SuperCM with 600 labels and varying values of $\beta$ on the CIFAR-10 dataset. Figure~\ref{fig:paramsen} shows the Top-1 accuracy of the model trained with different values of $\beta$, where $\beta=0$ represents the training with only supervised loss (CE). As CM is incorporated ($\beta >0$), we observe considerable performance improvement to existing CE baseline. However, we further observe that as the weight for the CM increases and less weight is given to the CE, performance slowly decreases.

\section{Conclusion}
\label{sec:conclusion}
In this paper, we present a simple end-to-end framework for SSL. Our training strategy benefits from the built-in clustering capability of the CM module and does not rely on complex training schemes. 
Facilitated by the differentiable CM, our method can be integrated into any gradient based SSL method as an unsupervised regularizer, paving the way to new versatile SSL approaches.


\end{document}